\documentclass{article}

\usepackage{arxiv}

\usepackage[utf8]{inputenc} 
\usepackage[T1]{fontenc}    
\usepackage[hidelinks]{hyperref}       
\usepackage{url}            
\usepackage{booktabs}       
\usepackage{amsfonts}       
\usepackage{nicefrac}       
\usepackage{microtype}      
\usepackage{lipsum}		
\usepackage{graphicx}
\usepackage[export]{adjustbox}
\graphicspath{ {./images/} }
\usepackage[ruled,vlined]{algorithm2e}
\usepackage[hidelinks]{hyperref} 
\usepackage{amsmath}

\usepackage{pgfplots}

\usepackage[font=small,labelfont=bf,tableposition=top]{caption}

\usepackage[colorinlistoftodos,prependcaption,textsize=tiny]{todonotes}

\usepackage[labelformat=simple]{subcaption}

\SetCommentSty{mycommfont}

\usepackage{comment}

\title{Privacy Preserving k-means clustering:\\A secure multi-party computation approach}

\date{\today} 					

\author{
	Daniel Hurtado Ram\'irez \\
	Department of Artificial Intelligence and Big Data\\
	GMV\\
	Isaac Newton,11, Tres Cantos,\\ Madrid, Spain \\
	\texttt{daniel.hurtado.ramirez@gmv.com} \\
	\And
	J. M. Au\~{n}\'{o}n \\
	Department of Artificial Intelligence and Big Data\\
	GMV\\
	Isaac Newton,11, Tres Cantos,\\ Madrid, Spain \\
	\texttt{jmaunon@gmv.com} \\
}

\renewcommand{\headeright}{}
\renewcommand{\undertitle}{}

\hypersetup{
pdftitle={Privacy Preserving k-means clustering:A secure multi-party computation approach},
pdfsubject={Privacy Preserving Machine Learning},
pdfauthor={Daniel Hurtado Ram\'irez, J.M. Au\~n\'on},
pdfkeywords={Privacy Preserving Machine Learning, Multi-party computation, K-means clustering},
}

\begin{document}
\maketitle

\begin{abstract}
	\label{sec:abstract}
	Knowledge discovery is one of the main goals of Artificial Intelligence. This Knowledge is usually stored in databases spread in different environments, being a tedious (or impossible) task to access and extract data from them. To this difficulty we must add that these datasources may contain private data, therefore the information can never leave the source. Privacy Preserving Machine Learning (PPML) helps to overcome this difficulty, employing cryptographic techniques, allowing knowledge discovery while ensuring data privacy. K-means is one of the data mining techniques used in order to discover knowledge, grouping data points in clusters that contain similar features. This paper focuses in Privacy Preserving Machine Learning applied to K-means using recent protocols from the field of criptography. The algorithm is applied to different scenarios where data may be distributed either horizontally or vertically.

\end{abstract}


\section{Introduction}
Privacy is one of society's greatest concerns. In 2018 the European Union published the \textit{General Data Protection Regulation} (GDPR)\cite{GDPR}, being this regulation \textit{an essential step to strengthen individuals' fundamental rights in the digital age and facilitate business by clarifying rules for companies and public bodies in the digital single market}. Nevertheless this concern comes from the origins of civilization. In the Roman Empire, Emperor Julius Caesar used cryptographic techniques (see Caesar ciphers \cite{CaesarCipher}) so that he could "share" his messages in a way that could not be revealed by his enemies.

In this paper the term "share" is closely related to Secure multi-party computation (SMPC), understanding SMPC as the set of techniques and protocols where several (multi) members (parties) wish to jointly compute (computation) a function $f(x_i)$ $\left(i=1,2,\dots p\right)$ using their entries, while keeping them private. SMPC has evolved a lot from its beginnings in the 80s, starting from theoretical research (see Yao and Shamir seminar papers \cite{Yao,ShamirShecret}), to real applications that allow the use of this technology to end users \cite{RealApp}. This evolution has been possible thanks to the effort of different research groups, which have tried to develop and optimize the protocols responsible of the computation (see \cite{MPCBook} and references therein)

On the other hand, in the last decade we have seen how the field of artificial intelligence has invaded our lives. We are surrounded by applications that use the data we generate in our day to day, sometimes in a controversial way \cite{NYTCambridge} but also helping society, such as the recent case of COVID-19 crisis managment \cite{NaturePeiffer2020machine}. These applications have one factor in common, they use data, and in many cases this data is private, therefore an entire area or research has emerged: Privacy Preserving Machine Learning (PPML) \cite{Surendra2017ARO,ppml2019}. In machine learning (and in statistics in general) one of the most common and fundamental problems is the clustering problem, understanding clustering as the way of partition data points into disjoint subsets, being these clusters represented by a set of features.

As the reader can infer after this short introduction, this paper will focus on the union of Privacy (via SMPC) and clustering, in particular the well-known K-means clustering algorithm \cite{benaloh1994dense,FUKUNAGA1990508}. This problem has been previously addressed depending on the distribution model (horizontal/vertical), the number of parties ($n\geq 2$) and the protocol itself \cite{KmeansVaidya, KmeansZhan,KmeansRafail, KmeansOrozco,KmeansSurvey}.

Here we do address, from our best of our knowledge, a novel way using additive secret sharing and SecureNN \cite{wagh2018securenn} protocols over horizontal and vertical private distributed data extended to $p$ parties.

The paper is organized as follows: Section \ref{sec:data-partitioning} briefly explains how data is typically distributed, explaining the difference between horizontal and vertical distributions; Section \ref{sec:k-means-centralized} introduces the K-means algorithm from a centralized point of view, illustrating the limitations and the conclusions that a user/researcher would obtain when lacking access to the data in a graphical way (Section \ref{sec:results}). Section \ref{sec:k-means-smpc} deals with the Secure version of the K-means algorithm, explaining the notation (Section \ref{sec:notation}) and the supporting protocols (Section \ref{sec:supporting-protocols}) which will be used (Section \ref{sec:main-protocols}).

\section{Data Partitioning}
\label{sec:data-partitioning}
The Abstract makes reference to the fact that data \textit{is usually stored in databases spread in different environments}. This information can have different structures, being partitioning commonly used when large datasets/tables are involved, so it is needed to split the information into smaller ones. In this paper partitioning will be used for representing how the dataset is split into different Parties $P_j$.

\begin{center}
	\begin{figure}[hb]
		\centering
		\captionsetup[subfigure]{justification=centering}
				
		\begin{subfigure}[t]{0.3\textwidth}
			\centering
			\includegraphics[page=1, width=0.75\linewidth]{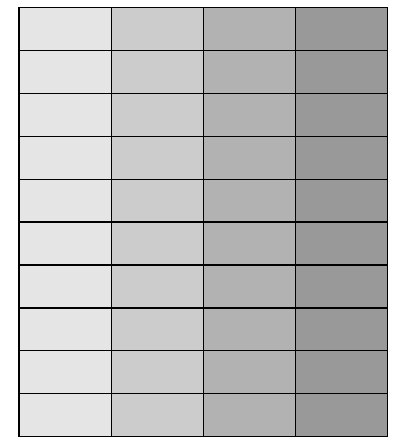}
			\caption{A single party has access to the whole dataset.}
			\label{fig:full_dataset }
		\end{subfigure}%
		\begin{subfigure}[t]{0.3\textwidth}
			\centering
			\includegraphics[page=2, width=0.75\linewidth]{../images/partition/partitioning.pdf}
			\caption{Horizontal partition. Several parties $P_j$ have different entries (rows), however all of them keep the same schema.}
			\label{fig:h_partition }
		\end{subfigure}%
		\begin{subfigure}[t]{0.4\textwidth}
			\centering
			\includegraphics[page=3, width=0.75\linewidth]{../images/partition/partitioning.pdf}
			\caption{Vertical partition. Several parties $P_j$ have different features (columns). There is no need to have the same schema.}
			\label{fig: v_partition }
		\end{subfigure}%
	\caption{Horizontal vs Vertical Partitioning}
	\label{fig:horizontal_vs_vertical}
	\end{figure}
		
\end{center}

Figure \ref{fig:horizontal_vs_vertical} depicts the different partitions we will work with. Figure \ref{fig:full_dataset } shows the case where all the data is centralized. This is the situation that is typically assumed when K-means is presented, and this is the one that will be assumed in Section \ref{sec:k-means-centralized}.

Figure \ref{fig:horizontal_vs_vertical} shows as well the situation where several parties have  either different rows (Figure \ref{fig:h_partition }) or columns (Figure \ref{fig: v_partition } ) of the same dataset. One of the main handicaps in Machine Learning is the lack of data to train, which leads to results not being accurate enough (however, getting more data can lead to privacy-related issues). Section \ref{sec:k-means-smpc}, and in particular, algorithms \ref{alg:H-kmeans}: \textsl{SHK-means}; and \ref{alg:V-kmeans}: \textsl{SVK-means} help to overcome this difficulty by using data from different parties but keeping its privacy. The advantages of having these algorithms will be illustrated in Section \ref{sec:results}. \ref{alg:V-kmeans}

\section{K-means algorithm: centralized approach}
\label{sec:k-means-centralized}
K-means algorithm is a well-known routine for finding clusters of points (represented by their centers) in an unlabeled dataset.
The usual K-means algorithm assumes that the we have full access to the data, leaving aside privacy concerns. In a non-formal language (formalism and notation will be introduced in Section \ref{sec:notation}), the steps are summarized in Algorithm \ref{alg:k-means} \cite{FUKUNAGA1990508}.

\begin{algorithm}[htb]
\SetAlgoLined
\SetKwInOut{Input}{Input}
\SetKwInOut{Output}{Output}
\Input{A matrix $X_{n \times d}$ of points, a matrix $C_{k \times d}$ of centers} 
\Output{Data points assigned to a cluster} 
\begin{enumerate}
    \item Select k random centers as the initial \textit{means}
\end{enumerate}
\Repeat{mean does not change}{
    \Indm
    \begin{enumerate}
        \setcounter{enumi}{1}
        \Indm 
        \item Calculate the distance between a data point and a center \tcp{Distance step}
        \item Assign each data point to the closest center \tcp{Labelling step}
        \item Re-calculate the center of each cluster perfoming the \textit{mean}
    \end{enumerate}}
 \caption{K-means}
 \label{alg:k-means}    
\end{algorithm}

The goal of this section is not to explain the algorithm itself (we invite to the reader to browse the bibliography for a detailed explanation), however core steps $2$ and $3$ will be briefly explained. In these steps the distance between a data point and a data center is calculated (step $2$), assigning a label depending on which cluster is closer (step $3$). Next sections will show how to adapt these main steps to SMPC framework, enabling the user to work with data that cannot be shared.

\subsection{Limitations of the centralized approach: Illustrated examples}
\label{sec:results}
In this section, we showcase the importance of Secure Multi Party Computation and K-means by providing two example situations in which it is essential. This section does not to demonstrate how good or bad the proposed algorithms will be, but rather to demonstrate how the conclusions obtained can change thanks to having access to private information in a secure way.

Addresed new algorithms in Section \ref{sec:main-protocols} have been developed on top of PySyft \cite{ryffel2018generic}. \textit{PySyft is a Python library for secure and private Deep Learning}, where most of the Additive Secret Sharing and secureNN \cite{wagh2018securenn} routines are implemented. This library (in particular PyGrid plattform) is also in charge of managing the communication between the parties involved in the computation. Thanks to this, we just need to focus on the computation itself, speeding up its implementation.

\begin{figure}[htb]
	\centering
	\includegraphics[width= {\textwidth},page=1]{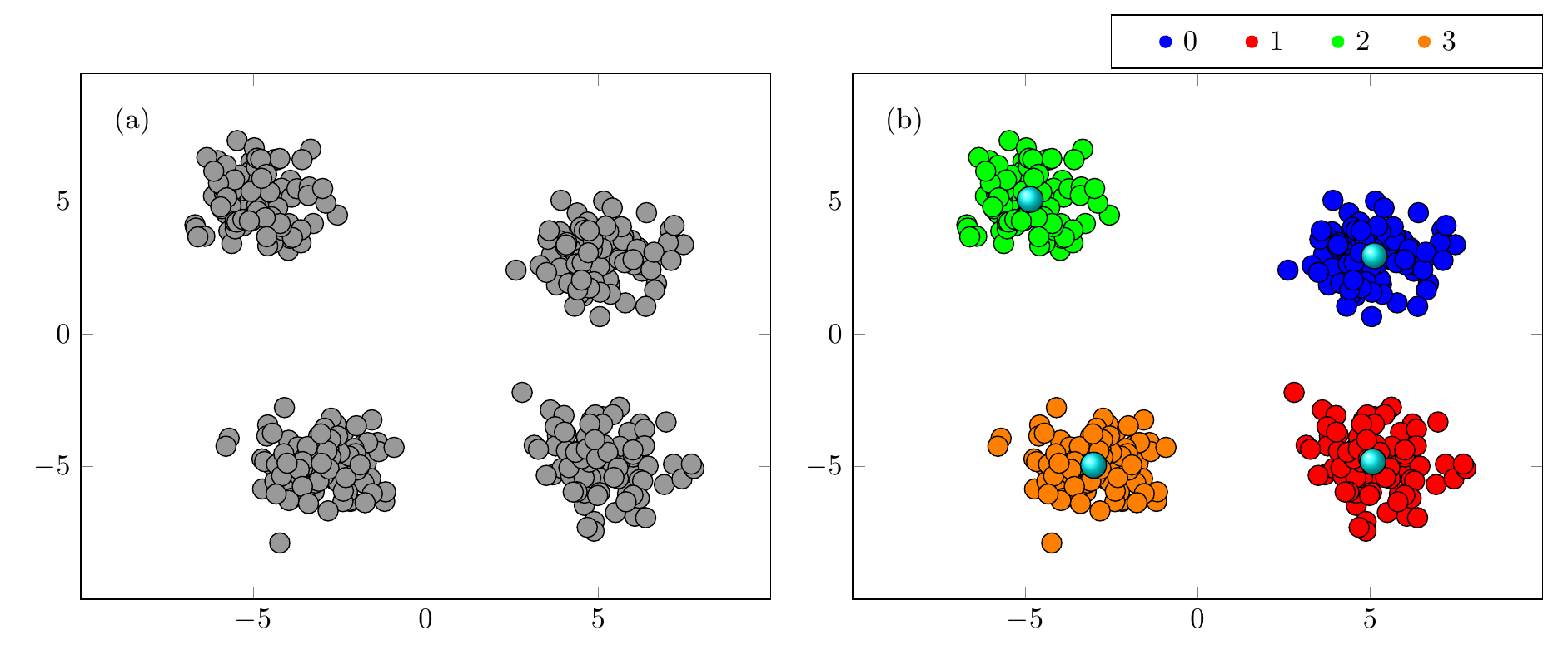}
	\caption{(a) Generated dataset $X$. (b) Labeled data points after applying K-means with $k=4$ to $X$. Centers $C$ are marked as cyan balls}
	\label{fig:results_1}
\end{figure}

For demonstration (and visualization) purposes we have generated a dummy dataset $X$ consisting of $4$ subsets of $100$ samples, where each subset follow a 2-dimensional normal Gaussian distribution of centers $\mu_0=(5,3)$, $\mu_1(5,-5)$, $\mu_2(-5,5)$ and $\mu_3(-3,-5)$; and a standard deviation $\sigma_0=\sigma_1=\sigma_2=\sigma_3=1$. Figure \ref{fig:results_1}(a) depicts the generated dataset that we call the \textit{Ground Truth}. 

We have imposed on the data $4$ clearly differentiated clusters. Figure \ref{fig:results_1}(b) shows the results of applying the centralized K-means algorithm (Algorithm \ref{alg:k-means}) with $k=4$ (this parameter will be constant throughout this entire subsection, assuming that a priori we don't know how many clusters we want to declare). As expected, we can see how the data points are assigned to one cluster or another depending on the distance to the centers, demonstrating how the algorithm works if it is trained with the whole dataset, nevertheless \textit{we do remark that Secure K-means algorithms can be considered as an alternative, obtaining exactly the same results, even in the case we couldn't access all the data}. These situations will be discussed in detail in the following subsections depending on data how data is distributed.

\subsection{Horizontal distribution}
\begin{figure}[htb]
	\centering
	\includegraphics[width= {\textwidth},page=2]{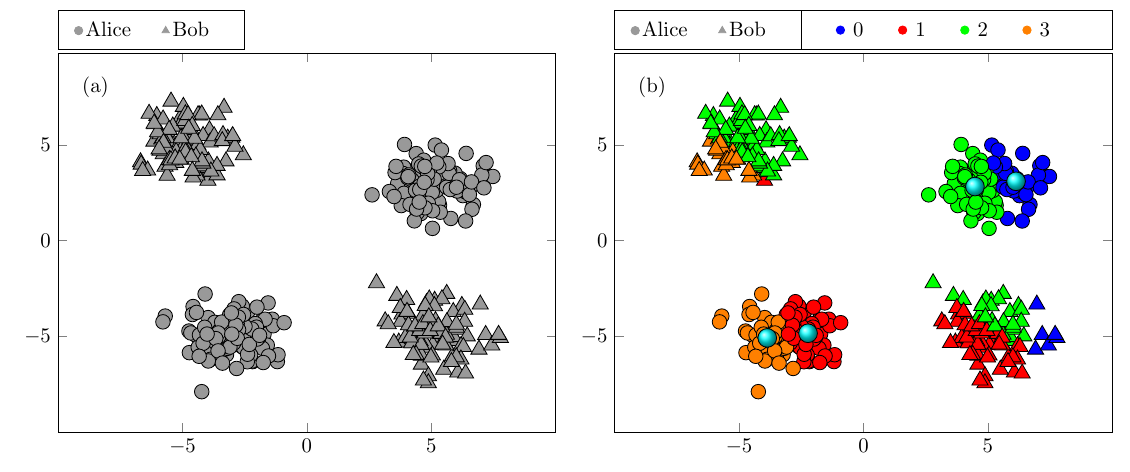}
	\caption{(a) Dataset distribution between Alice and Bob. (b) Labeled data points after  training the centralized K-means ($k=4$) with data from Alice ($\circ$) and testing with data from Bob ($\bigtriangleup$). Centers $C$ from training phase are marked as cyan balls.}
	\label{fig:results_2}
\end{figure}

Now we address the situation where data is horizontally distributed between the fictional famous   characters Alice and Bob \cite{Rivest78amethod}, namely, $X_{400\times 2}=X^{(A)}_{200\times 2} \cup X^{(B)}_{200\times 2}$, being the intersection between $X^{(A)}$ and $X^{(B)}$ an empty set ($X^{(A)}\cap X^{(B)}=\emptyset$). Figure \ref{fig:results_2}(a) shows this distribution in a graphical way. Notice that the difference between both comes from the samples, the number of columns remains the same.

Now Alice wants to train  the centralized \textsl{K-means} algorithm with her data, so she will train the algorithm:
\begin{itemize}
	\item \textit{Training phase}. $X_{train}=X^{(A)}$. The output of this  step is the centers $C$.
\end{itemize}

For the testing phase, Alice and Bob reach an agreement and Alice sends the model (centers in this case) to Bob, then Bob just have to analyze the results with his own data (data never seen for the algorithm):
\begin{itemize}
	\item \textit{Testing phase}. Bob has centers $C$, so the testing dataset will be $X_{test}=X^{(B)}$. Bob calculates the distance between $C$ and $X_{test}$.
\end{itemize}

Figure \ref{fig:results_2}(b) shows theses steps in a single snapshot. The \textit{training phase} ($\circ$: Alice's data) shows that points have been assigned to $4$ clusters which obviously are not the original ones that \textit{Ground truth} illustrates in Figure \ref{fig:results_1}(a). The centers $C$ (marked as cyan balls) are placed into different position with respect Figure \ref{fig:results_1}(b) and for that reason in the \textit{testing phase} ($\bigtriangleup$: Bob's data) the new points are assigned to clusters to which they do not really belong. In a real use case, the conclusions reached could be totally wrong, due to the algorithm having been trained without sufficient data.

Notice that Alice and Bob could have obtained exactly the same results as the centralized version (Figure \ref{fig:results_1}(b)) but without sending any private information, if they  had additively shared their data and had executed  Algorithm \ref{alg:H-kmeans}: \textsl{SHK-means}. In this way, Alice could learn nothing from Bob but centers $C$.

\subsection{Vertical distribution}

\begin{figure}[htb]
	\centering
	\includegraphics[width= {\textwidth},page=3]{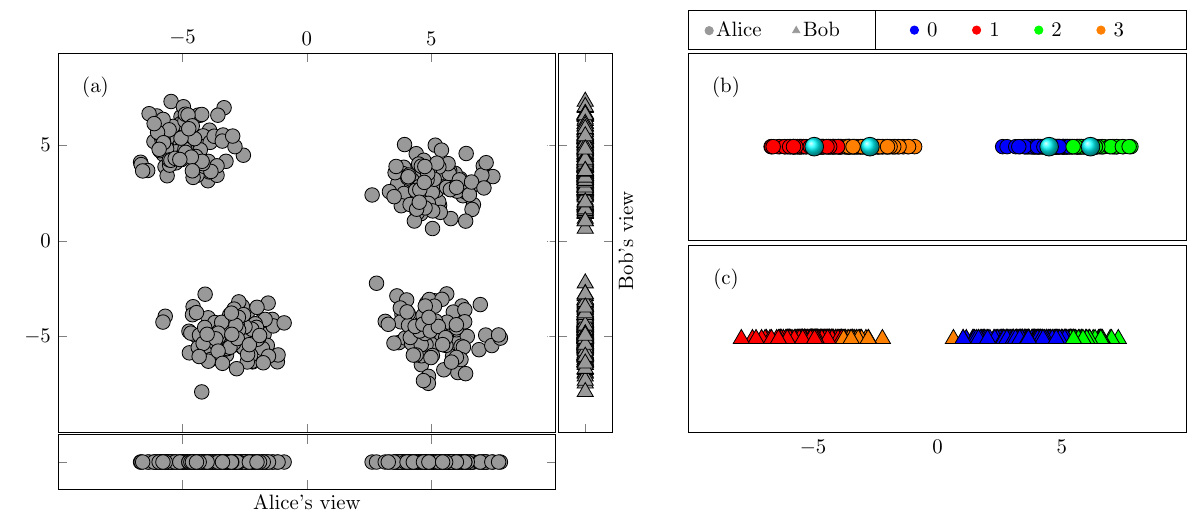}
	\caption{(a) Dataset distribution between Alice and Bob. Alice and Bob have a 1-dimensional dataset, then points are plotted showed in a straight line (Alice's view (bottom) and Bob's view (right)) (b) Labeled data points after  training the centralized K-means ($k=4$) with data from Alice ($\circ$) and (c) testing with data from Bob ($\bigtriangleup$). Centers $C$ from training phase are marked as cyan balls.}
	\label{fig:results_3}
\end{figure}

In this subsection data will be vertically distributed between Alice and Bob, i.e., $X_{400\times 2}=X^{(A)}_{400\times 1} \cup X^{(B)}_{400\times 1}$ (satisfying $X^{(A)}\cap X^{(B)}=\emptyset$). This situation can be more striking because instead of having a 2-dimensional array (points belong to a plane), each party has a 1-dimensional array (points belong to a straight line), so they watch the \textit{Ground truth} from a different perspective. Figure \ref{fig:results_3}(a) shows this situation, attaching the views from Alice and Bob perspective's for clarification purposes.

Now the same procedure as in the previous subsection is followed: Alice trains the centralized K-means algorithm with her own data ($X^{(A)}$), sends the model to Bob, and Bob tests the model with his own data ($X^{(B)}$). Figure \ref{fig:results_3}(b) shows the training performed by Alice with their respective centers. The algorithm has tried to find 4 clusters along the straight line, clusters that obviously do not correspond to the original ones. Figure \ref{fig:results_3}(c) shows the testing performed by Bob. In this case, whereas Bob has 2 main clusters, the points have been assigned according to the centers $C$ sent by Alice, mixing the real clusters with the ones assigned by the algorithm. Therefore, we can again say that the conclusions reached could be totally wrong, due to the algorithm having been trained without the proper features.

In order to obtain the same conclusions that they would obtain as if they had access to whole data, Alice and Bob can use Algorithm \ref{alg:V-kmeans}: \textsl{SVK-Means}, In this way, none of them would send any original data but they would get the same advantages as if they had it.

\section{K-means algorithm: SMPC approach}\label{sec:k-means-smpc}
\subsection{Notation}
\label{sec:notation}
Previous Section has explained the centralized K-means and the issues one can find if has no access to data. To overcome these difficulties we will focus on  privacy issues. This section establishes the notation that will be used in the following sections. From \cite{wagh2018securenn} we borrow the notation and the following protocols: \textsl{secure matrix multiplication, DReLU} and \textsl{division}. Note that \textsl{DReLU} stands for Derivative of \textsl{ReLU}; that is, $\textsl{DReLU}(x) = 0$ if $x < 0$ and $\textsl{DReLU(x)} = 1$ if $x > 0$ 

We wish to allow an arbitrary number of parties to provide data and take part in the computation. The protocols in SecureNN are presented for the 2-party case, but they can be easily extended to work with $p$ parties; we provide an example of such an adaptation in our \textsl{ElementWiseMatMul} protocol, an adaptation of SecureNN's \textsl{MatMul}. The protocols require the presence of an extra party that assists in the computations without providing data; we take $P_0$ as the assistant party, i.e., the responsible of generating triplets for Beaver multiplication \cite{Beaver}, and $P_1, P_2, \dots, P_p$ to be the parties holding the data.

We assume the aggregate of all parties' data (which we will call $X$) consists of $n$ samples of $d$-dimensional data, and this data is shared additively across parties $P_1, P_2, \dots, P_p$. $\langle X \rangle_j$ refers to $P_j$'s share of $X$, such that $X = \sum_{j=1}^p \langle X \rangle_j$ (over $\mathbb{Z}_L$, a finite field with size $L$ ). $X[i]$ refers to sample $i$, $X[i][j]$ refers to coordinate $j$ of sample $i$ ($1 \leq i \leq n$, $1 \leq j \leq d$).

The goal is to split the data into $k$ clusters. These clusters are defined by their centers $C$, where $C[j]$ refers to the center of cluster $j$ ($1 \leq j \leq k$). The centers will also be shared across all parties, so $C = \sum_{j=1}^p \langle C \rangle_j$

We wish to reveal as little data as possible; this includes the original samples, the resulting clusters, the label of any given sample, and the distance between the samples and the cluster centers. We consider acceptable revealing how many samples correspond to each cluster (but not which ones) since nothing useful can be extracted from this information while providing a significant speedup of the algorithm; however, the algorithm can easily be modified to keep this information private.

\subsection{Supporting protocols}
\label{sec:supporting-protocols}
Now we present some preliminary protocols, which will be used as building blocks by the main algorithms presented in Section \ref{sec:main-protocols}.

\subsubsection{Element-wise Matrix Multiplication}
One of the core operations for the secure K-means algorithm is multiplication. Algorithm \ref{ElementWiseMatMul}: \textsl{ElementWiseMatMul} addresses the elementwise secure matrix multiplication, i.e., $Z_{n\times d}=X_{n\times d} \odot Y_{n \times d}$ ($\odot$ symbol representing the elementwise multiplication), where parties $P_j$ $(j\ge 1)$ hold shares of $X,Y$. This algorithm is the elementwise version of $\Pi_{\textrm{MatMul}}$ (see Reference \cite{wagh2018securenn}), extended to $p$ parties.

\begin{algorithm}[H]
	\SetAlgoLined
	\SetKwInOut{Input}{Input}\SetKwInOut{Output}{Output}
	\Input{Shared matrices $X_{m \times n}$ and $Y_{m \times n}$; that is, $P_j$ holds $\left(\langle X\rangle_j, \langle Y \rangle_j\right)$ for all $j \in \{1, 2, \dots p\}$}
	\Output{A shared matrix $Z_{m \times n}$ such that $Z[i][j] = X[i][j] \cdot Y[i][j]$}
	\begin{enumerate}
		\item $P_0$ computes $\langle u \rangle_1, \dots, \langle u \rangle_p$, shares of zero matrices of size $m \times n$ (that is, $\sum_{l=1}^{p} \langle u \rangle_l = 0^{m \times n}$) and sends $\langle u \rangle_i$ to $P_i$, $1 \leq i \leq p$
		\item $P_0$ picks random matrices (over $\mathbb{Z}_L$) $A_{m \times n}$ and $B_{m \times n}$, computes $C_{m \times n} = A \odot B$
		\item $P_0$ generates shares for $A, B, C$ and sends them to $P_1, P_2, \dots, P_p$
		\item For $j \in \{1, 2, \dots, p\}$, $P_j$ computes $\langle E \rangle_j = \langle X \rangle_j - \langle A \rangle_j$ and $\langle F \rangle_j = \langle Y \rangle_j - \langle B \rangle_j$
		\item $P_1, P_2, \dots, P_p$ reconstruct $E$ and $F$ by exchanging shares
		\item For $j \in \{2, \dots, p\}$, $P_j$ outputs $Z_j :=  \langle u \rangle_j + \langle X \rangle_j \odot F +  \langle Y \rangle_j \odot E +  \langle C \rangle_j$; \\
		      $P_1$ outputs $Z_1 := \langle u \rangle_1 + \langle X \rangle_1 \odot F + \langle Y \rangle_1 \odot E +  \langle C \rangle_1 - E \odot F$
	\end{enumerate}
	\caption{ElementWiseMatMul}
	\label{ElementWiseMatMul}
\end{algorithm}

\subsubsection{Distance Matrix}
In Algorithm \ref{alg:k-means} the \texttt{Distance step} was described as one of the fundamental steps of K-means. Algorithm \ref{MatDist}: \textsl{MatDist} computes the secure squared euclidean distance $d^2$ between two shared vectores $\mathbf{x}$, $\mathbf{y}$: $d^2\left(\mathbf{x},\mathbf{y}\right)$=$\left( x_1 -  y_1\right)^2 + \left( x_2 -  y_2\right)^2 + \cdots + \left( x_d -  y_d\right)^2$= $\sum_{j=1}^{d}\left( x_j -  y_j\right)^2$, namely between a data point $X\left[i\right]$ and a center $C\left[j\right]$.

\begin{algorithm}[H]
	\SetAlgoLined
	\SetKwInOut{Input}{Input}
	\SetKwInOut{Output}{Output}
	\Input{A shared matrix $X_{n \times d}$ of points, a shared matrix $C_{k \times d}$ of centers}
	\Output{A shared matrix $M_{n\times k}$ where $M[i][j]$ is the euclidean squared distance from point $X[i]$ to center $C[j]$ }
	\begin{enumerate}
		\item Each party $p$ reshapes $\langle X \rangle_p^{n \times d}$ into $\langle A \rangle_p^{n \times k \times d}$, such that $\langle A \rangle_p[i][l][j] = \langle X \rangle_p[i][j]$ for all $i, j, l$ (with $l$ being a free index, $1 \leq l \leq k$)
		\item Each party $p$ reshapes $\langle C \rangle_p^{k \times d}$ into $\langle B \rangle_p^{n \times k \times d}$, such that $\langle B \rangle_p[l][i][j] = \langle C \rangle_p[i][j]$ for all $i, j, l$ (with $l$ being a free index, $1 \leq l \leq n$)
		\item Each party $p$ computes $\langle D\rangle_p = \langle A \rangle_p - \langle B \rangle_p$. Intuitively, $D[i][j][l]$ is the distance in the dimension $l$ between center $i$ and cluster $j$
		\item Use \textsl{ElementWiseMatMul} to square all elements of $D$
		\item Each party $p$ outputs $\langle M\rangle_p$, with $\langle M\rangle_p[i][j] := \sum_{l=1}^d \langle D\rangle_p[i][j][l]$
	\end{enumerate}
	\caption{MatDist}
	\label{MatDist}
\end{algorithm}

\subsubsection{Labelling}
Last supporting algorithm is Algorithm \ref{LabelSamples}: \textsl{LabelSamples}. It addresses the "label" of a data point $X\left[i\right]$, meaning that a point will belong to a cluster if and only if the distance between this point and the $j^{th}-$cluster is closer than the rest. This algorithm is the secure version of  \texttt{Labelling step} from Algorithm \ref{alg:k-means}

\begin{algorithm}[tbh]
	\SetAlgoLined
	\SetKwInOut{Input}{Input}
	\SetKwInOut{Output}{Output}
	\Input{A shared matrix $X_{n \times d}$ of points, a shared matrix $C_{k \times d}$ of centers}
	\Output{A shared matrix $H_{n\times k}$ with the labels of samples $X$, where $H[i][j]$ is 1 if and only if center $j$ is the closest to sample $i$}
	\begin{enumerate}
		\item Use \textsl{MatDist} to compute $M := \textsl{MatDist}\left(X, C\right)$ in a secure manner
		\item For all $l \in \{1, \dots, k\}$, compute $H_l[i][j] := DReLU\left(M[i][l] - M[i][j]\right)$. Intuitively, $H_l[i][j]$ is 1 if and only if cluster $j$ is closer to point $i$ than cluster $l$
		\item Use \textsl{ElementWiseMatMul} to compute $H[i][j] := \prod_{l=1}^{k}H_l[i][j]$. We have that $H[i][j] = 1$ if and only if $H_l[i][j] = 1$ for all $l$, that is, cluster $j$ is the closest cluster to point $i$
		\item Return $H$
	\end{enumerate}
	\caption{LabelSamples}
	\label{LabelSamples}
\end{algorithm}

\subsection{Main protocols}
\label{sec:main-protocols}
\subsubsection{Secure Horizontal K-means}
Figure \ref{fig:h_partition } showed the case where data is horizontal partitioned into different parties. \textsl{SHK-means} deals with this situation, translating the core steps from Algorithm \ref{alg:k-means} into the SMPC framework. It takes all the data as an additively shared matrix, which allows the data to be arbitrarily distributed between all parties.

\begin{algorithm}[tbh]
	\SetAlgoLined
	\SetKwInOut{Input}{Input}\SetKwInOut{Output}{Output}
	\Input{A shared matrix $X_{n \times d}$ of points, the public number $k$ of clusters, a public number $\epsilon > 0$ for the stopping criterion}
	\Output{A shared matrix $C_{k \times d}$ of cluster centers}
	\begin{enumerate}
		\item Select random public integers $l_1, l_2, \dots, l_k$, $1 \leq l_j \leq n$. Each party $p$ sets $C_p[j] = X_p[l_j]$, for all $j, 1 \leq j \leq k$
	\end{enumerate}
	\While{true}{
		\begin{enumerate}
			\setcounter{enumi}{1}
			\Indm
			\item Compute $H := \textsl{LabelSamples}(X, C)$ using the \textsl{LabelSamples} protocol
			\item Set $t_j := \sum_{i=1}^n H[i][j]$, the total number of samples that go to cluster $j$
			\item Set $T_{k \times d} := \textsl{MatMul}\left(H^T, X\right)$. Intuitively, row $j$ of $T$ is the sum of all samples that belong in cluster $j$
			\item Compute the new centers as $\tilde{C}[j] := T[j]/t_j$. This division can either be done with a secure division protocol (completely secure, but slow) or by first revealing the values $t_j$ to both parties
			\item For each $j \in \{1, \dots, k\}$, use the $MatDist$ protocol to compute $D_j = \textsl{MatDist}\left(C[j], \tilde{C}[j]\right)$
			\item Set $C := \tilde{C}$
			\item Compute the additively shared value $\Delta = \sum_{j=1}^k D_j$, the total movement of all centers
			\item Compute $s := \textsl{DReLU}\left(\epsilon - \Delta\right)$ and reconstruct its value. If $s = 1$, stop and return $C$; if $s = 0$, return to step 2
		\end{enumerate}
	}
	\caption{Secure Horizontal K-means (SHK-means)}
	\label{alg:H-kmeans}
\end{algorithm}

This algorithm learns how data points are grouped while keeping privacy. Taking advantange of Machine Learning terminology, this phase is usually denoted as \textit{training phase}. In the \textit{testing phase} (or prediction), i.e., when Centers are already determined and new data points $Y$ (additively shared across parties) need to be labeled, it is enough to compute $H = \textsl{LabelSamples}(Y, C)$ and reconstruct it. Thanks to the fact that $H$ is a one-hot encoded shared matrix, it could also be directly used as an input to some other secure protocol that builds on top of K-means, without losing any privacy.

We are also aware of communication costs (this will be detailed in Section \ref{sec:communication}), for that reason, and as alternative, we suggest to reveal $t_j$, that is, the total number of data points which belong to $j^{th}$-cluster. This improves the performance of the algorithm.

\subsubsection{Secure Vertical K-means}

The second algorithm is fine tuned for the special case in which the data is Vertically Partitioned across parties (see Figure \ref{fig: v_partition }); that is, each party has a different dimension of some common entities. This allows for most of the computation to be done locally, which greatly improves performance. For simplicity purpouses, we slightly modify the notation, assuming that party $P_j$ has column $j$ of the data. Each party will end up with column $j$ of the cluster centers. Figure \ref{fig:vertical_X} tries to clarify to  to the reader to this point.

\begin{figure}[htb]
	\begin{equation*}
		X=\left[\begin{array}{ccccccccccc}
				X_{1} & \mid & X_{2} & \mid & \cdots & \mid & X_{j} & \mid & \cdots & \mid & X_{p}\end{array}\right]
	\end{equation*}
	\caption{Notation used in SVK-means algorithm. It is remarked with vertical bars that two columns $X_i$, $X_j$ are hold by different parties. Same applies to Centers $C$}
	\label{fig:vertical_X}
\end{figure}

\begin{algorithm}[htb]
	\SetAlgoLined
	\SetKwInOut{Input}{Input}\SetKwInOut{Output}{Output}
	\Input{The data columns $X_1, X_2, \dots X_p$ (each belonging to a different party), the public number $k$ of clusters, a public number $\epsilon > 0$ for the stopping criterion}
	\Output{The centroids' columns $C_1, C_2, \dots C_p$ (each belonging to a different party)}
	\begin{enumerate}
		\item Select random public integers $l_1, l_2, \dots, l_k$, $1 \leq l_j \leq n$. Each party $p$ sets $C_p[j] = X_p[l_j]$, for all $j, 1 \leq j \leq k$
	\end{enumerate}
	\While{true}{
		\begin{enumerate}
			\setcounter{enumi}{1}
			\Indm
			\item $P_0$ computes $\langle u \rangle_1, \dots, \langle u \rangle_p$, shares of zero matrices of size $n \times k$ (that is, $\sum_{l=1}^{p} \langle u \rangle_l = 0^{n \times k}$) and sends $\langle u \rangle_i$ to $P_i$, $1 \leq i \leq p$
			\item Compute the additively shared matrix $D_{n \times k}$, by having each party $p$ set $\langle D \rangle_p[i][j] := \langle u \rangle_p + (X_p[i] - C_p[j])^2$ (a local version of \textsl{MatDist})
			\item Compute $E[i] = \textsl{ArgMin}_{j \in \{1, 2, \dots, k\}}(D[i][j])$, the closest cluster center for each point
			\item $P_1, \dots, P_p$ reconstruct $E$ by exchanging shares
			\item For each $j \in \{1, \dots, k\}$, set $t_j = \left|\{ i \in \{1, \dots, n\} \; | \; E[i] == j\}\right|$, the total number of points that are assigned to cluster $j$
			\item For each $j \in \{1, \dots, k\}$ each party $p$ sets $\tilde{C}_p[j] = \left( \sum \{ X_p[i] \; | \; i \in \{1, \dots, n\}, E[i] == j\} \right) / t_j$
			\item Compute the additively shared value $\Delta$, by having each party $p$ set $\langle \Delta \rangle_p = \sum_{j=1}^{k} (C_p[j] - \tilde{C}_p[j])^2$
			\item Each party $p$ sets $C_p := \tilde{C}_p$
			\item Compute $s := \textsl{DReLU}(\epsilon - \Delta)$ and reconstruct its value. If $s = 1$, stop and have each party return $C_j$; if $s = 0$, return to step 2
		\end{enumerate}
	}
	\caption{Secure Vertical K-means (SVK-means)}
	
	\label{alg:V-kmeans}
\end{algorithm}

Note that Algorithm \ref{alg:V-kmeans} uses the protocol \textsl{ArgMin}, which securely computes the index of the minimum value of a set of values. Such a protocol can be obtained as a small adaptation of SecureNN's \cite{wagh2018securenn} \textsl{Maxpool} protocol.

To label new samples (\textit{predict}), it is enough to perform steps 2 and 3 using the new samples $Y$ instead of $X$.

\subsection{Communication complexity analysis}
\label{sec:communication}

The performance bottleneck in all of the presented protocols is in the communication rounds that they require. In this section, we study how these costs scale with the number of clusters $k$ (the number of samples and their dimension only affect the size of the messages, not their amount). Table \ref{table:communication} summarizes this information.

	\begin{minipage}[c]{.5\textwidth}
		\centering
		\begin{tabular}{ll}
			\toprule
			Protocol                    & Rounds                   \\
			\midrule
			\textsl{ElementWiseMatMul}  & 2                        \\
			\textsl{MatDist}            & 2                        \\
			\textsl{LabelSamples}       & 2k + 8                   \\
			SHK-means (secure division) & 2k + 150 (per iteration) \\
			SHK-means (fast division)   & 2k + 20 (per iteration)  \\
			SVK-means                  & 9k                       \\
			\bottomrule
		\end{tabular}
		\captionof{table}{Communication rounds for each protocol}
		\label{table:communication}
	\end{minipage}%
	\begin{minipage}[c]{.5\textwidth}
		\centering
		\includegraphics[width=0.75\textwidth]{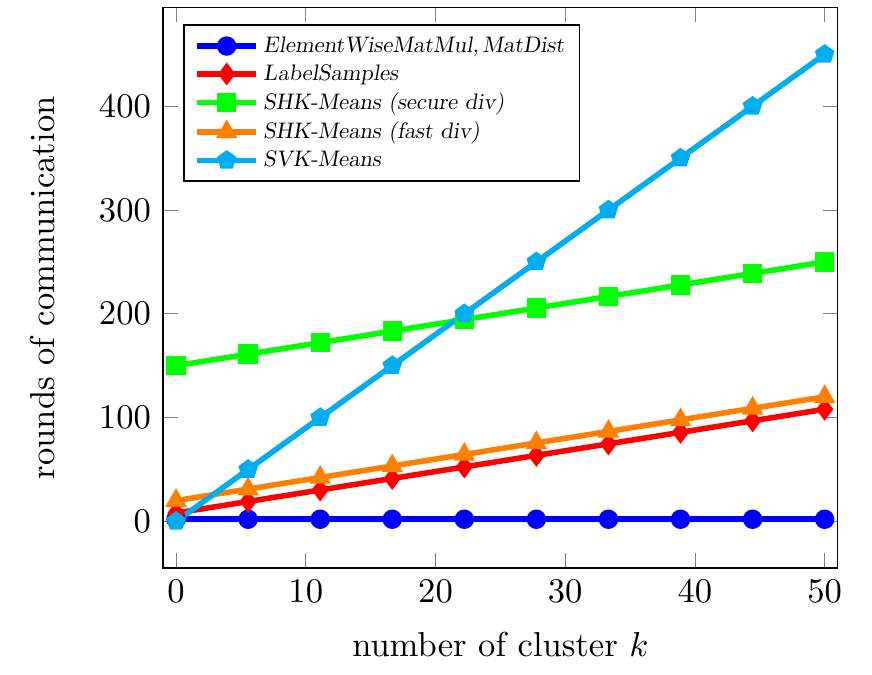}
		\captionof{figure}{Scalability of the algorithms with the number of clusters}
		\label{fig:communication}
	\end{minipage}

The \textsl{MatDist} protocol requires communication only in the 4th step, a call to \textsl{ElementWiseMatMul}, which requires 2 rounds of communication.

The \textsl{LabelSamples} protocol requires $2k + 8$ rounds of communication: 2 from using the \textsl{MatDist} protocol, 8 from the \textsl{DRelu} protocol (if we vectorize the computation of all the $H_l$'s into a single batch), and $2(k-1)$ from using $k-1$ times the \textsl{MatMul} protocol.

Each iteration of the \textsl{Horizontal K-means} protocol requires either $2k + 20$ or $2k + 20 + 130$ rounds of communication: $2k + 8$ from using the \textsl{LabelSamples} protocol, $2$ from the multiplications in step 4 (vectorized appropiately), either $130$ or 0 rounds for step 5 (depending on the choice to perform the division securely or not), and 10 rounds to compute the stopping criterion.

Each iteration of the \textsl{Secure Vertical K-means} protocol requires $9k$ rounds of communication: $9(k - 1)$ from using the \textsl{ArgMin} protocol in step 3, one round for reconstructing $E$ in step 4, and 8 rounds to compute the stopping criterion.

Note that the \textsl{SVK-means} protocol will require more communication rounds than the \textsl{SHK-means} protocol for $k > 2$, but the messages are considerably smaller, which makes it faster in practice. Figure \ref{fig:communication} depicts the rounds of communication versus the number of cluster for every used algorithm.

Protocols and algorithms addressed use SecureNN as a base; nevertheless, it can be replaced by any other secure computation framework that supports the basic operations of addition, multiplication, comparison, and division. We do cite FALCON \cite{wagh2020falcon} or AriaNN \cite{ryffel2020ariann} as recent examples that could be used.

\section{Conclusions}
\label{sec:conclusions}
K-means clustering is a basic and essential tool for data scientists. We have presented secure
versions of this algorithm, allowing the use of previously inaccessible data, and showing how not using secure algorithms, and thus
lacking access to the full dataset, can lead to wrong models and conclusions. These algorithms
can be used by any number of parties, and they are tailored both for horizontal and vertical 
data distributions. 

With this work, we have also shown a general approach for adapting traditional machine learning algorithms to a secure setting.

\section{Acknowledgements}
\label{sec:acknowledgments}
We thank the OpenMined community, whose efforts have provided a solid base on which to experiment and iterate.
This work is partially supported by Spain's Ministerio de Economía y Empresa (TSI-100906-2019-2) and GMV.

\bibliographystyle{unsrt}

\end{document}